\documentclass[a4paper,11pt journal,twocolumn]{IEEEtran}
\IEEEoverridecommandlockouts
\usepackage{cite}
\usepackage{amsmath,amssymb,amsfonts}
\usepackage{graphicx}
\usepackage{textcomp}

\usepackage{url}

\usepackage{caption}
\usepackage{multirow}

\usepackage{enumerate}
\let\labelindent\relax
\usepackage{enumitem}
 \usepackage{algorithm}
 \usepackage[algo2e,ruled,lined]{algorithm2e}

 \usepackage{algpseudocode}
 \usepackage{algcompatible}
 \usepackage{graphicx}
 \usepackage{multirow}
 \usepackage[table,xcdraw]{xcolor}
 \usepackage[colorlinks]{hyperref}
\usepackage{amsmath}
\usepackage{rotating}
\usepackage{colortbl}

\usepackage{array}
\usepackage{float}
\newcolumntype{A}{>{\centering\arraybackslash}m{1.2 cm}}
\newcolumntype{C}{>{\centering\arraybackslash}m{1.32 cm}}
\newcolumntype{E}{>{\centering\arraybackslash}m{1.4 cm}}

\def\BibTeX{{\rm B\kern-.05em{\sc i\kern-.025em b}\kern-.08em
    T\kern-.1667em\lower.7ex\hbox{E}\kern-.125emX}}
\begin{document}

\title{Novel Human Machine Interface via Robust Hand Gesture Recognition System using Channel Pruned YOLOv5s Model \\
{\footnotesize \textsuperscript{}}
\thanks{}
}

\author{Abir~Sen,~\IEEEmembership{Student Member,~IEEE}
        ~Tapas~Kumar~Mishra,~\IEEEmembership{Member,~IEEE}
        and ~Ratnakar~Dash,~\IEEEmembership{Member,~IEEE}
}

\maketitle

\begin{abstract}
Hand gesture recognition (HGR) is a vital component in enhancing the human-computer interaction experience, particularly in multimedia applications, such as virtual reality, gaming, smart home automation systems, etc. Users can control and navigate through these applications seamlessly by accurately detecting and recognizing gestures. However, in a real-time scenario, the performance of the gesture recognition system is sometimes affected due to the presence of complex background, low-light illumination, occlusion problems, etc. Another issue is building a fast and robust gesture-controlled human-computer interface (HCI) in the real-time scenario. The overall objective of this paper is to develop an efficient hand gesture detection and classification model using a channel-pruned YOLOv5-small model and utilize the model to build a gesture-controlled HCI with a quick response time (in ms) and higher detection speed (in fps). First, the YOLOv5s model is chosen for the gesture detection task. Next, the model is simplified by using a channel-pruned algorithm. After that, the pruned model is further fine-tuned to ensure detection efficiency. We have compared our suggested scheme with other state-of-the-art works, and it is observed that our model has shown superior results in terms of mAP (mean average precision), precision (\%), recall (\%), and F1-score (\%), fast inference time (in ms), and detection speed (in fps). Our proposed method paves the way for deploying a pruned YOLOv5s model for a real-time gesture-command-based HCI to control some applications, such as the VLC media player, Spotify player, etc., using correctly classified gesture commands in real-time scenarios. The average detection speed of our proposed system has reached more than 60 frames per second (fps) in real-time, which meets the perfect requirement in real-time application control. We also depicted the real-time performances of the gesture-controlled VLC and Spotify player with respect to the detection rate (\%) and response time (in ms) and improved the robustness of our gesture-controlled applications.

\end{abstract}


\section{Introduction}

Human-computer interaction (HCI) is the key to the success of interactive systems. It involves combining knowledge of human abilities with technical knowledge of hardware and software technologies.
Recently hand gesture recognition has become an integral part of human-computer interaction (HCI). Hand gesture-controlled HCI provides a highly flexible and user-friendly environment to enable the connection between man and machine without physically touching them. There are several modalities for human-machine interfaces, such as graphical user interface (GUI), voice-command-based, hand-gesture controlled, touch-screen-based, etc. Among those modalities, hand gesture recognition has emerged as a promising approach due to its natural and more convenient nature. Robust hand gesture recognition systems play a crucial role in facilitating effective human-machine communication and improving the user experience. Consider the following example: (1) One user wishes to utilise gesture commands to operate the smart TV. For instance, using the gestures Five/Fist in front of the TV to turn it on or off, or Hang/L to adjust the smart TV's volume, etc. Therefore, in the aforementioned scenarios, one user can operate the smart TV using only correct gesture commands without pressing any remote buttons, which is advantageous for those who require physical assistance to use the applications. The two types of hand gesture recognition systems are wearable sensor-based\cite{berezhnoy2018hand,abhishek2016glove} and vision-based \cite{singha2018dynamic,al2022structured} recognition systems. The disadvantage of the sensor-based recognition system is that the user must wear the glove and requires a wired connection to classify the gestures. But sometimes, it produces false gesture classes when there is a complex background. The pipeline of a vision-based gesture-recognition system consists of four stages, (1) capturing images from a web camera, (2) segmentation to remove background, (3) feature extraction, and classification of gesture images. Yet, difficulties arise when developing a gesture recognition system in a chaotic or complicated context. Over the past decade, hand gesture recognition has become vital to facilitating human-machine interaction as it eliminates the physical dependency on a keyboard and mouse. However,  building a robust hand gesture recognition system in the presence of a cluttered background and varying light conditions, is a challenging task. To address these issues, this study develops a robust hand gesture recognition based on lightweight channel-pruned YOLOv5s model and design of an HMI that utilises predicted gesture class labels as input commands to control multimedia. The significant contributions of our work are summarized as follows:
\begin{enumerate}[label=\Roman*]

    \item Development of a gesture recognition system using fine-tuned YOLOv5s model under complex background and light varying conditions.
    
    \item To enhance the gesture detection model performance and reduce the inference time, the fine-tuned YOLOv5s model is pruned using a channel pruning algorithm.

    \item Development of an HMI to control VLC and Spotify player with quick response time (in ms) and higher inference speed (in fps) by utilizing the fine-tuned pruned YOLOv5s model file.

\end{enumerate}

\hspace{-0.6cm} The remainder of this article is structured as follows. Section \ref{literature} contains reviews of the literature pertaining to various hand gesture methodologies. Section \ref{proposed} covers the proposed approaches. Section \ref{experiment} 
describes dataset description, experimental details, comparative analysis, and a real-time performance study. The discussion section is covered in Section \ref{discussion}. Section \ref{conclusion} discusses the conclusion of the study and outlines potential future work.

\section{\label{literature} Literature survey}

 The advancement of computer vision in recent years has greatly facilitated human-machine interaction through hand gesture detection. Computer vision researchers utilize webcams and hand gestures as the main inputs for gesture recognition systems. However, operating multimedia applications with gesture commands presents obstacles such as gesture segmentation, cluttered backgrounds, and low-light conditions. So, recognizing gestures is a primitive task for developing a gesture-controlled HMI. Second, we require a lightweight deep learning model that can accurately detect and recognize hand gestures, map them, and use them as gesture commands to control multimedia applications in real-time with quick response time. Various state-of-the-art works for hand gesture recognition have been reported in the last decades. Some conventional approaches for vision-based gesture classification are Hidden Markov model (HMM), K-nearest neighbor algorithms, and support vector machine (SVM). 
For instance, Elmezain \textit{et al.} \cite{elmezain2008hidden} have proposed an automatic system to recognize both isolated and continuous gestures for Arabic numbers 0 to 9 from stereo color image sequences based on the motion trajectory of a single hand using HMM and Gaussian mixture model (GMM). In \cite{yang2012dynamic}, the authors have suggested a gesture recognition system that consists of four phases: (1) skin color-based segmentation, (2) the use of a spotting algorithm to find out the starting and ending of a gesture portion, (3) extraction of some features such as hand position, velocity, shape followed by combining all features to obtain feature vector, (4) classification using HMM. A static hand gesture recognition system was proposed by Huang \textit{et al.} \cite{huang2009vision}. They have used the Gabor filter for feature extraction tasks and SVM classifier classification. Their proposed approach has achieved a recognition rate of 95.2\% and a processing time of 0.2 sec per frame. Another contribution of  Rahman and Afrin \cite{rahman2013hand} presents a static hand gesture recognition system using SVM classifier. In this work, the Canny edge detector is used to detect the edges of a gesture portion, followed by extracting features using bi-orthogonal wavelet transformation. Next, the SVM classifier is employed to classify the gestures. But recently, with the advent of deep learning, the CNN algorithm has gained popularity for extracting high-level features and classification tasks. Many state-of-the-art works have been reported regarding hand gesture classification using CNN. For instance,  Yingxin \textit{et al.} \cite{yingxin2016robust} have developed a hand gesture recognition system utilising canny edge detection to detect the gesture edge in pre-processing phase and convolutional neural network (CNN) for feature extraction. Ultimately, they employed a fully connected dense layer to obtain the final class label. Sharma \textit{et al.} \cite{SHARMA2021115657} have suggested a deep learning approach after modifying VGG11 and VGG16 architecture to classify static hand gestures of sign language. In their experiment, they have considered ASL and ISL datasets to evaluate the model performances. They have achieved the classification accuracy of 94.84\% and 99.96\%. \color{black} 
Sen \textit{et al.} \cite{sen2022novel} have devised a novel system for recognizing hand gestures. The system consisted of three stages: (1) detecting gestures using binary thresholding, (2) segmenting the gesture portion, and (3) training three custom CNN classifiers simultaneously. The output scores of these models were then calculated to construct an ensemble model for the final prediction.

The difficulty still lies in developing a robust framework for gesture recognition that yields precise outcomes. However, it faces obstacles in real-world scenarios, including lighting conditions, complex backgrounds, varying distances, and occlusion issues. The researchers find it difficult to detect and recognize the correct gesture classes under these conditions with quick response time and higher detection speed. Under these scenarios, object detection algorithms such as, YOLO \cite{redmon2016you}, Faster-RCNN (region-based convolutional neural networks) \cite{girshick2015fast} play a vital role in localizing and recognizing the gesture classes in real-time. There are some research articles regarding the robust hand gesture recognition system using object detection algorithms. For instance, Bose \textit{et al.} \cite{bose2020efficient} have
Introduced a hand gesture recognition system in the presence of intricate backgrounds. They utilized the Faster-RCNN framework with the Inception-V2 model as feature extractors. Their proposed method has achieved a detection time of 140ms per gesture. In \cite{yu2019hand}, the authors have proposed a gesture recognition system utilizing the Faster RCNN model. 
They train the model using the Standard hand gesture database under complicated backgrounds. Their classification accuracy exceeds 90.6\%. Recently, sen \textit{et al} \cite{sen2024hgr} has suggested a robust hand gesture recognition system based on frozen YOLOv5s model under complicated backgrounds and low-light conditions. In their work, they have optimized the YOLOv5s architecture by freezing some convolutional layers in the backbone portion, resulting in a reduction in the number of parameters, model size, and inference time. Their proposed framework is evaluated on two annotated hand gesture datasets: one private (`NITR-HGR') and one public (`ASL'). In their work, they have created the `NITR-HGR' dataset with five individuals (four men and one woman) from 'NIT Rourkela'. Their suggested framework has achieved a mean average precision (mAP@50-95) above 92\%, along with a detection speed of 55 frames per second (fps).
In recent years, gesture-controlled human-machine interfaces (HMIs) have attracted a great deal of interest, particularly in virtual reality, gaming, and automotive interfaces. However, 
constructing the interfaces is challenging because of the uninterrupted succession of frame sequences in the real-time video stream. Only a handful of works on low-cost gesture-controlled HMI are available. For instance, in \cite{rautaray2010novel}, the authors developed a gesture-controlled HCI to operate a media player. Four stages comprise their proposed scheme: (1) portion detection of the hand; (2) center finding of the hand using the K-mean algorithm; (3) generating a rectangle around the hand's center and classifying it using the K-nearest neighbor algorithm; and (4) utilizing the predicted gesture-commands to control different functions of a media player. Recently, in \cite{sen2023deep}, the authors have shown a gesture recognition system based on CNN models and a vision transformer. Moreover, they have depicted a gesture-controlled HMI to operate VLC, Spotify, and 2D-Mario-Bros game in real-time by using the correctly classified gesture classes obtained from the best model (among pre-trained CNN models and vision transformer).
A few issues with their suggested approach include poor background, lower recognition rate (\%), a longer inference time, a lack of resilience, etc. So, this work suggests a robust hand gesture recognition system using an optimized YOLOv5s model and a gesture-controlled HMI that can map the recognized gestures to put commands/specific actions to control multimedia applications with lower latency and higher detection speed. 

 \section{\label{proposed} Suggested channel-pruned YOLOv5s for gesture classification}
\begin{figure*}[!ht]
  \centering
  \includegraphics[width=\textwidth,height=7 cm]{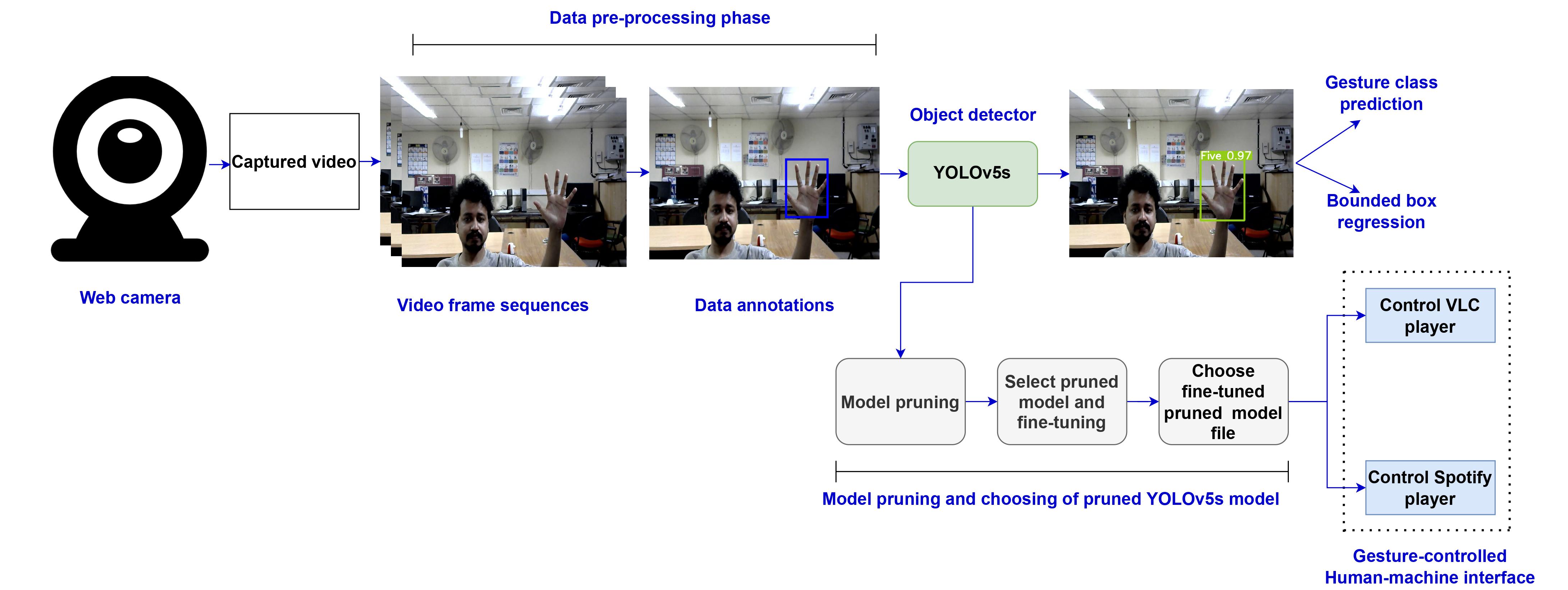}
  \caption{\label{model_diagram_4th} Block diagram of our suggested scheme.}
  \end{figure*}

This section details our proposed scheme for developing a novel HMI via a robust hand gesture recognition system using a channel-pruned YOLOv5s model to operate multimedia applications in real-time scenarios. Our proposed methodology consists of five vital stages: (1) data acquisition, (2) pre-processing phase including data annotation, (3) utilization of the YOLOv5 model for both detection and classification tasks, (4) pruning of the YOLOv5s model to reduce the number of parameters and model size (in MB), (5) utilize the lightweight pruned model to design a gesture-controlled HMI in real-time scenarios. The block diagram of this framework is depicted in Figure \ref{model_diagram_4th}, and the steps followed to build the channel pruned YOLOv5s model are illustrated in Algorithm \ref{algo:pruned_yolov5s}. All the stages of building the model framework are described as follows:
\begin{algorithm}
  	\caption{Suggested channel-pruned YOLOv5s (small version) model}\label{algo:pruned_yolov5s}
  		\textbf{INPUT}: Load the trained YOLOv5s model (trained on labeled dataset).
  		 
  		\textbf{OUTPUT}: Obtain the lightweight pruned YOLOv5s model.
            
            Let's assume the scaling factors of channels $c_{1}$, $c_{2}$,  $...c_{n}$,  are $\gamma_{1}$, $\gamma_{2}$,  $...\gamma_{n}$. \\
       \textbf{Initialize}: epoch $\gets 1$,  epochs $\gets 60$, pruning rate $\gets 0.20$ \\
            Start the sparse training with a sparsity rate.\\
             \While{epoch $\leq$ epochs}{
                        
                    \textbf{Step 1:} Sort the scaling factors ($\gamma$) values of BN layers in ascending order to find the pruning threshold.
                    \\
                      $sl$ $\gets$ sorted list of $\gamma$ values.
                    
                    \textbf{Step 2:} Calculate the highest pruning threshold by calculating the minimum of the maximum gamma values of each BN layer to ensure that not all channels will be pruned.
                    \\
                    \textbf{Step 3:} Find the threshold index for pruning to determine the position in the sorted list of scaling factors ($\gamma$) at which channels will be pruned.

                     threshold index $\gets$ len($sl$) $\times$ pruning rate 

                    \textbf{Step 4:} Compute the pruning threshold based on the threshold index.
                    
                     pruning threshold $\gets$ $sl$[threshold index]

                    \While{$sl$ $\neq$ NULL}
                    {
                    
                    $f$ $\gets$ $sl$[0] (choose the first $\gamma$ value from the sorted list and compare it with the pruning threshold.)\\
                         \If { $f$  $<$ pruning \ threshold}{

                         The channels whose scaling factors ($\gamma_{1}$, $\gamma_{2}$,  $...\gamma_{n}$) are less than pruning threshold, will be eliminated.

                          $sl$ $\gets$ $sl$ $\setminus f$ 
                        
        }
        
}

     }

  Return the channel pruned YOLOv5s model.    
\end{algorithm}

\subsection {Gesture detection based on YOLOv5-small as baseline model}\label{gesture_detect_4th_yolo}
YOLO \cite{redmon2016you} is a one-stage network for object recognition that transforms object classification into a regression problem. 
Among the four variants of YOLOv5 models (small, mega, large, and extra large), 
 YOLOv5-small (YOLOv5s) is considered the fastest and performs best on two popular object detection
datasets, Pascal VOC \cite{everingham2009the} and Microsoft COCO \cite{lin2014microsoft} datasets. 
There are three component parts that make up the YOLOv5s model: (1) the backbone, (2) the neck, and (3) the detection section. A brief description of each component is provided below.
\paragraph{\textbf{Backbone layer}} 
Three components make up the backbone layer \cite{redmon2016you}\cite{sen2024hgr}: (1) the focus layer, (2) CSPDarkNet, and (3) SPP. The backbone network of YOLOv5 is built upon CSPDarknet53, which is a modified version of Darknet53 featuring a 'cross-stage partial network' (CSP) architecture aimed at enhancing the data transmission capabilities of the network.

\begin{itemize}
    \item \textbf{Focus layer} It is used to reduce the computational complexity of the model by minimizing the feature map's spatial dimensions and the number of parameters. 
    
    \item \textbf{SPP layer} SPP \cite{he2015spatial} stands for spatial pyramidal pooling layer. It is used after one or more convolutional layers. Its objective is to enable predictions at multiple spatial resolutions during a single forward pass for the model.

\end{itemize}

\paragraph{\textbf{Neck layer}}The FPN and PAN structures are used to combine features from different levels of the backbone via bottom-up and top-down pathways in order to improve the distribution of semantic features and locate data. It aids in aggregating and enhancing features in order to capture information at various scales, which is essential for object detection.

\paragraph{\textbf{Detection layer}}
The detection layer in YOLOv5s estimates the likelihood of each object belonging to a specific class. The YOLOv5s model comprises three detection layers, each corresponding to one of the three processing scales employed for the input image. The final output is generated by combining the results of these three detection layers through concatenation.

In our experiment, the YOLOv5s model was initialized with the model parameters, which were pre-trained on the COCO dataset \cite{lin2014microsoft}.
The gesture images are scaled to 480 × 480 to fine-tune the YOLOv5s-based gesture detection model. The learning rate, batch size, and epochs are set to 0.01, 32, and 60. SGD optimizer is used to optimize the network parameters iteratively.

\subsection{Channel-Pruned YOLOv5s model}\label{channel_pruning}

\begin{figure*}
\centering
\includegraphics[width=16 cm,height=3 cm]{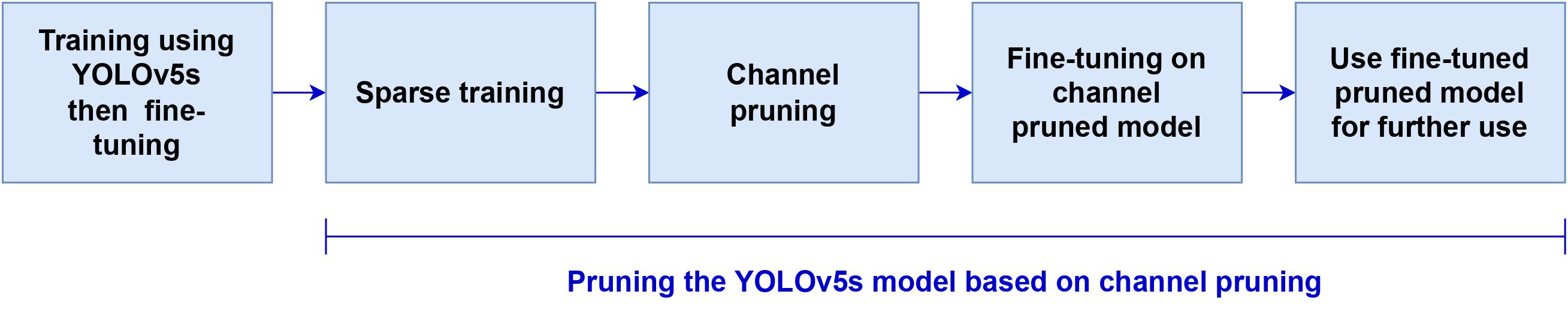}
\caption{\label{pruning_4th} Pipeline of channel-pruned YOLOv5s model.}
\end{figure*}

   In the channel pruning algorithm, the batch normalization (BN) layer is introduced with the scaling factor (denoted as $\gamma$). After training both scaling factors and weight parameters simultaneously, the scaling factors ($\gamma$ values) are sorted, followed by calculating the pruning threshold. The channels whose corresponding scaling factors are below the pruning threshold values will be eliminated. Finally, the pruned model becomes lightweight and fine-tuned to improve the performance. The main stages of channel pruning are (i) sparsity training, (ii) channel pruning, and (iii) fine-tuning of the pruned model. The pipeline of channel-pruned-based YOLOv5s model is shown in Figure \ref{pruning_4th}. The implementation steps to build the proposed channel-pruned YOLOv5s model are shown in Algorithm \ref{algo:pruned_yolov5s}, and the details of every stage are described below.   
\subsubsection*{\textbf{Sparse training}}
 During sparse training, the scaling factor of the BN layers of the YOLOV5s model was subjected to the L1 norm to make the model more sparse. During sparse training, both scaling factors and weight parameters are simultaneously trained. After sparse training, the values of $\gamma$ are approaching 0, indicating that the associated channels with smaller scaling factors will be eliminated.  In sparse training, the choice of the loss function depends on the objective of the training and the technique being used to induce sparsity in the model.
The loss function \cite{liu2017learning} during the sparse training is given by the following Equation \ref{cost_4th}.
\begin{equation}\label{cost_4th}
  Loss = \sum l(g(x,w),y) + \lambda \sum k(\gamma) 
\end{equation}
Here $x$ and $y$ denotes the training input and output parameters and w represents the weight parameter.\\ $\lambda \sum k(\gamma) $ term denotes the penalty term that is added to the loss function. In this term $\lambda$ denotes the penalty factor and $k(\gamma) $ represents the sparsity penalty on the scaling factor $\gamma$. 
\\
\subsubsection*{\textbf{Channel pruning} }
\begin{figure}
\centering
\includegraphics[width=9cm,height=5.8cm]{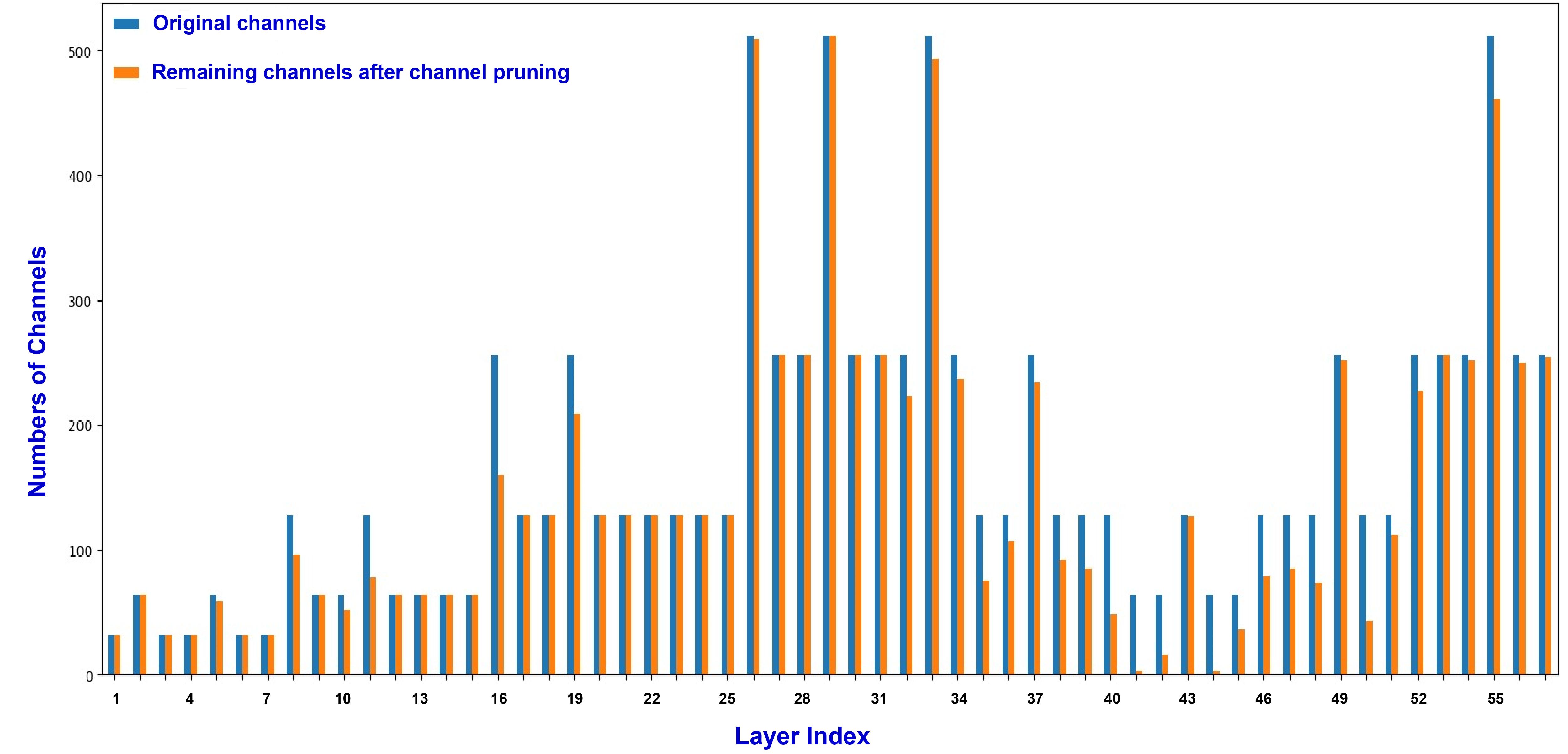}
\caption{\label{pruning_channel} Layer index-wise changes of channels, before and after channel pruning after using pruning rate 15\%.}
\end{figure}

After sparse training, the $\gamma$ values from the BN layers are sorted to find the threshold value. The next phase is to identify the position in the sorted list of BN weights at which the channels will be pruned using the pruning threshold index. Following this, the pruning threshold is calculated utilizing the threshold index. Both the formulas to calculate the threshold index and pruning threshold are shown in Equations \ref{thindex} and \ref{pruning_th}. 

\begin{equation}\label{thindex}
     threshold \ index= length (sb) \times pruning \ rate
\end{equation}
\begin{equation}\label{pruning_th}
    pruning \ threshold = sb[threshold \ index]
\end{equation}
where sb denotes the sorted list of BN weights and pruning rate specifies what proportion of the channels will be pruned.\\
The channels whose scaling factors ($\gamma$) are less than the pruning threshold, will be eliminated. Subsequently a compact channel-pruned model is obtained with less number of parameters and computation cost. \\
In the experiment, the pruning rate has been set from 10\% to 20\% as higher pruning rates potentially degrade model performance due to the reduction of the number of trainable parameters.
We have compared the performance of YOLOv5s model on the NITR-HGR dataset at pruning rates varying from 10\% to 20\% and the comparison results (after fine-tuning of channel pruned YOLOv5s models) have been shown in Table \ref{compare_prune_baseline}. The original channels and remaining channels after using the channel pruning method are shown in Figure \ref{pruning_channel}.

\subsubsection* {\textbf{Fine-tuning of the pruned model}}
The pruning rate reduces the detection accuracy after channel pruning due to reducing the number of channels. So, the channel-pruned model is further trained with the hyper-parameter settings to enhance the detection performance. Like the training phase, in the fine-tuning phase, the batch size and number of epochs are both configured to 32 and 60, respectively. The SGD optimizer is employed to update the model parameters throughout the training process.

\section{\label{experiment} Experimental Evaluation}
To evaluate the effectiveness of the suggested framework, simulations have been carried
out on two annotated hand gesture datasets: one private dataset, `NITR-hand gesture' (NITR-HGR) \cite{sen2024hgr} (mentioned in our previously published article \cite{sen2024hgr}), and `American sign language' (ASL) (public dataset) datasets. The data collection and annotation details have been explored in subsection \ref{dataset_details}.

\subsection{Dataset collection and details}\label{dataset_details}
 The NITR-hand gesture dataset\cite{sen2024hgr} was created with the assistance of five individuals from `NIT Rourkela' (four men and one woman). Initially, the recordings were captured using the web camera at a frame rate ranging from 25 to 30 fps. The frames were subsequently extracted from the videos.
 Consequently, the bounding box annotates the gesture region within the frames. There are thirteen different classes (example, Three, Two, Thumb, Palm etc.) of gesture images in this dataset. There are 250 images in each gesture class, each having a resolution of $(640 \times 480)$ pixels. 
  \\

The ASL dataset consists of 29 gesture classes, including 26 for the letters A through Z, and 3 additional classes. We have collected 2844 annotated gesture samples from the Roboflow \cite{asl-recognition-uhnrr_dataset}.
\subsection{\label{experiment_details}Details of the experimental setup}

To carry out our experiment, the data set is split with a ratio of 80:20 for training and testing. 
All the experiments were carried out on a linux-based system that has a 16GB NVIDIA graphics card, 64GB RAM. The input images are reduced to an aspect ratio of (480 × 480) during the training phase. Subsequently, the batch size and number of epochs are adjusted to 32 and 60.

\subsection{Results \label{result}}
In this section, we have displayed various experimental results evaluated on the `ASL' and `NITR-HGR' datasets. Four evaluation metrics, including precision (P), recall (R), F-score values (F1-score), and mean average precision (mAP), are selected in order to assess the detection performance. Equations \ref{PR} to \ref{mAP} show the precision, recall, F-score, and mAP formulas. 

\begin{equation}\label{PR}
    precision=\frac{TP}{TP+FP} 
\end{equation}
\begin{equation}\label{RE}
recall=\frac{TP}{TP+FN}
\end{equation}

\begin{equation}\label{F-score}
F-score=\frac{2\times PR \times RE}{PR+RE}
\end{equation}

\begin{equation}\label{mAP}
mAP = \frac{1}{\textit{N}} \sum_{\textit{i}=1}^{\textit{N}} \text{average(precision)}_\textit{i}
\end{equation}


\subsubsection{\textbf{Results on NITR-HGR dataset}}\label{results_nitr_hgr}

\begin{figure*}[!ht]
\centering
\includegraphics[width=12cm,height=4.2 cm]{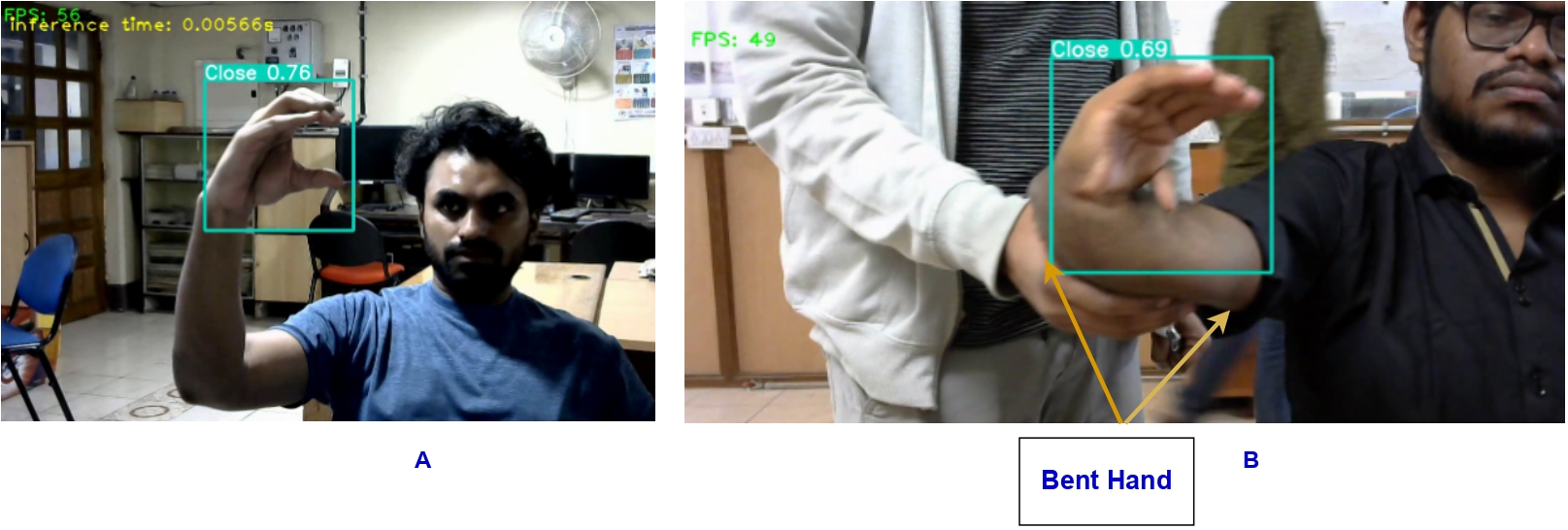}
\caption{\label{correctly_classified_nitr} Illustration of two correctly classified gesture samples using the fine-tuned channel pruned YOLOv5s model, (A) normal person, (B) physically impaired person (with bend hand). }
\end{figure*}

 To assess the efficacy of the channel pruned-YOLOv5s-based gesture detection model, it is compared with the YOLOv5s baseline model on the NITR-HGR dataset. The performance analysis is shown in Table \ref{compare_prune_baseline}. 
  Figure \ref{correctly_classified_nitr} exhibits two correctly classified gesture samples for physically impaired and normal individuals by using a 20\% fine-tuned pruned YOLOv5s model. 
 
\begin{table*}[!ht]
    
\centering
\caption{\label{compare_prune_baseline} Performance comparison of YOLOv5s model for gesture detection after fine-tuning of channel pruned models with different pruning rates (\%) on NITR-HGR dataset.}
\label{compare_prune_baseline}
\resizebox*{0.90\textwidth}{!}{ 
\begin{tabular}{|ccccccccc|}
\hline
\multicolumn{1}{|c|}{\textbf{Models}}              & \multicolumn{1}{c|}{\textbf{\begin{tabular}[c]{@{}c@{}}Pruning \\ rate (\%)\end{tabular}}} & \multicolumn{1}{c|}{\textbf{\begin{tabular}[c]{@{}c@{}}Number of\\ parameters\\ (in millions)\end{tabular}}} & \multicolumn{1}{c|}{\textbf{GFlops}} & \multicolumn{1}{c|}{\textbf{\begin{tabular}[c]{@{}c@{}}Model\\ size (MB)\end{tabular}}} & \multicolumn{1}{c|}{\textbf{F-score}} & \multicolumn{1}{c|}{\textbf{mAP 50-95}} & \multicolumn{1}{c|}{\textbf{\begin{tabular}[c]{@{}c@{}}Inference \\ time (in ms)\end{tabular}}} & \textbf{\begin{tabular}[c]{@{}c@{}}Average detection\\ speed (in fps)\end{tabular}} \\ \hline
\multicolumn{1}{|c|}{Baseline YOLOv5s}             & \multicolumn{1}{c|}{0}                                                                     & \multicolumn{1}{c|}{7.05}                                                                                    & \multicolumn{1}{c|}{15.1}            & \multicolumn{1}{c|}{14.5}                                                               & \multicolumn{1}{c|}{99.65}            & \multicolumn{1}{c|}{92.50}              & \multicolumn{1}{c|}{7.1}                                                                        & 54                                                                                  \\ \hline
\multicolumn{9}{|c|}{\textbf{Suggested Pruned YOLOv5s model (after fine-tuning)}}                                                                                                                                                                                                                                                                                                                                                                                                                                                                                                                                                                                         \\ \hline
\multicolumn{1}{|c|}{\textbf{10\%-pruned YOLOv5s}} & \multicolumn{1}{c|}{10}                                                                    & \multicolumn{1}{c|}{6.2}                                                                                     & \multicolumn{1}{c|}{13.1}            & \multicolumn{1}{c|}{12}                                                                 & \multicolumn{1}{c|}{99.65}            & \multicolumn{1}{c|}{92.50}              & \multicolumn{1}{c|}{6.2}                                                                        & 59                                                                                  \\ \hline
\multicolumn{1}{|c|}{\textbf{15\%-pruned YOLOv5s}} & \multicolumn{1}{c|}{15}                                                                    & \multicolumn{1}{c|}{5.8}                                                                                     & \multicolumn{1}{c|}{12.1}            & \multicolumn{1}{c|}{11.5}                                                               & \multicolumn{1}{c|}{99.60}            & \multicolumn{1}{c|}{91.10}              & \multicolumn{1}{c|}{6.1}                                                                        & 60                                                                                  \\ \hline
\multicolumn{1}{|c|}{\textbf{20\%-pruned YOLOv5s}} & \multicolumn{1}{c|}{20}                                                                    & \multicolumn{1}{c|}{5.3}                                                                                     & \multicolumn{1}{c|}{11.2}            & \multicolumn{1}{c|}{10.7}                                                               & \multicolumn{1}{c|}{99.00}            & \multicolumn{1}{c|}{90.80}              & \multicolumn{1}{c|}{6.0}                                                                        & 62                                                                                  \\ \hline
\end{tabular}

}
\end{table*}

\subsubsection*{\textbf{Model performance analysis of channel pruned YOLOv5s model after fine-tuning}} 
Table \ref{compare_prune_baseline} demonstrates the performance comparison of the baseline YOLOv5s model and the pruned YOLOv5s models after fine-tuning on the validation set of NITR-HGR dataset at different pruning rates (\%). Specifically, the model parameters, FLOPs, and the model size (MB) are used to compare the efficiency of various models. FLOPs are used to quantify the computational complexity of models. In Table \ref{compare_prune_baseline}, it is noticed that the GFLOPS and the number of parameters are reduced after using different pruning rates (\%). It is observed in Table \ref{compare_prune_baseline} that 
the pruning rate of 10\% to 20\% resulted in a reduction of the number of GFLOPS in pruned YOLOv5s models by 13\%, 18\%, and 25\% respectively, as compared to the baseline YOLOv5s model. In case of inference time, it has been reduced by 0.9 to 1.1 ms by using a different pruning rate.
Similarly, the model size is reduced by increasing the pruning rate (\%). When compared to the baseline YOLOv5s model, the fine-tuned pruned model performs well in terms of inference time (in ms) and detection speed (in fps), despite a  reduction in detection accuracy (in mAP).

\subsubsection*{\textbf{Comparative analysis with different object detection models on NITR-HGR dataset}}
To further analyze the performance of the suggested model, 12 object detection models were evaluated on the NITR-HGR dataset. These models included Faster RCNN (with Inception-V2 as the backbone), SSD, YOLOv3, YOLOv5s (with ghost module), YOLOv5s (with transformer module), two larger versions of the YOLOv5 model such as YOLOv5-mega, YOLOv5-large, YOLOv7 and four frozen YOLOv5s models \cite{sen2024hgr} (after freezing different layers). Table \ref{model_comparison_nitr} depicts the comparison analysis of the suggested model with 12 different models in terms of evaluation metrics. It is observed in Table \ref{model_comparison_nitr}, that the precision, recall, F1-score, and mAP values of our suggested framework are higher than the other state-of-the-art works presented by \cite{rubin2019hand}, \cite{mujahid2021real}, \cite{hu2022gesture} and \cite{sen2024hgr}. It is also noticed in Table \ref{model_comparison_nitr} that both YOLOv7 and our suggested fine-tuned pruned YOLOv5s (with pruning rate 10\%) have the highest precision compared to other models, but its size is larger than our suggested framework. Table \ref{model_comparison_nitr} shows that the size of the YOLOv5s model with ghost module is lowest compared to other models, but the precision, recall, F1-score, and mAP of this model are less than YOLOv5s (with transformer module), YOLOv7 and suggested fine-tuned pruned YOLOv5s model. Therefore, based on the comparison analysis, it can be concluded that the proposed framework performs better in detecting and classifying gestures in low-light conditions and complex backgrounds when compared to the other eight object detection models.

\subsubsection{ \textbf{Results on ASL dataset}}\label{results_asl_dataset}
This subsection provides an overview of the experiments performed on the ASL dataset in order to address the 29-gesture class problem. The experiments were performed using our proposed model on the ASL dataset, with the default input size of ($224 \times 224$).
 Five object detection models were trained on the labeled ASL dataset. These models are Faster RCNN (with Inception-V2 as the backbone), three larger versions of the YOLOv5 model such as YOLOv5-mega, YOLOv5-large, YOLOv5-extralarge, and YOLOv7. It is noticed in Table \ref{model_comparison_asl_4th}, that the precision of YOLOv5l is larger than other used models but less than our suggested 10\% pruned YOLOv5s model. Table \ref{model_comparison_asl_4th} shows that the performance of the suggested model has outperformed the other scheme presented by \cite{rubin2019hand} and the four used object detection models. 

\begin{table*}[!ht]
\centering
\caption{\label{model_comparison_nitr} 
Experimental results of different object detection-based models on NITR-HGR dataset.}
\label{model_comparison_nitr}
\resizebox*{0.90\textwidth}{!}{
\begin{tabular}{|c|c|c|c|c|c|c|c|} 
\hline
\textbf{Models}                                                                   & \textbf{Precision (\%)} & \textbf{Recall (\%)} & \textbf{F-score (\%)} & \textbf{mAP50 (\%)} & \begin{tabular}[c]{@{}c@{}}\textbf{mAP}\\ \textbf{50-95 (\%)} \end{tabular} & \begin{tabular}[c]{@{}c@{}}\textbf{Parameters} \\ (\textbf{in millions})  \end{tabular} &  \begin{tabular}[c]{@{}c@{}}\textbf{Model} \\ \textbf{size (in MB)}  \end{tabular} \\ 
\hline
\begin{tabular}[c]{@{}c@{}}Faster RCNN\\ (Inception V2) \cite{rubin2019hand} \end{tabular}     & 88.32          & 87.60       & 87.95        &  90.45     &  84.60                                                        &    50.62    & 135                                                              \\ 
\hline
\begin{tabular}[c]{@{}c@{}}SSD \\ $(320 \times 320)$\end{tabular}        &  90.52              &  89.65           &  90.08            &    92.50        &   84.65                                                       &     15.7      & 44                                                       \\ 

\hline
\begin{tabular}[c]{@{}c@{}}YOLOv3  \cite{mujahid2021real}\\ (DarkNet53 as backbone)  \end{tabular}        & 94.82           & 95.12        & 94.97        & 90.80       & 84.10                                                     & 62.5   & 235                                                               \\ 

\hline
\begin{tabular}[c]{@{}c@{}}YOLOv5s \\ (Ghost module) \cite{hu2022gesture}\end{tabular}        & 98.6           & 98.20        & 98.40         & 92.80       & 84.20                                                     & 3.7  & 9.40                                                                 \\ 
\hline
\begin{tabular}[c]{@{}c@{}}YOLOv5s \\ (Transformer module)\end{tabular}  &  99.60              &    99.50         &  99.55            &        99.40    &    91.06                                                      &    7.05    & 15                                                              \\ 
\hline
YOLOv5m                                                                  &  96.40         &   95.70          &      96.05        &   98.4         &    85.32                                                      &  21   & 40.7                                                                 \\ 
\hline
YOLOv5l                                                                  & 95.24               &     94.12        &   94.68           &     97.67       &   84.46                                                       &     46       & 92.7                                                          \\ 
\hline
YOLOv7                                                                   &    99.60            &    98.40         &    99.00          &    99.40        &    92.45                                                      &    37.2       & 74.9                                                           \\ 
\hline

\hline
\begin{tabular}[c]{@{}c@{}}YOLOv5s\\ after freezing 3 layers \cite{sen2024hgr}\end{tabular}                      &             99.50                            & 99.35                                       &   99.42                                          &            99.26                              &                        92.60 &    7.01      & 13.80                                                                    \\
\hline
\begin{tabular}[c]{@{}c@{}}YOLOv5s \\  after freezing 5 layers \cite{sen2024hgr}\end{tabular}                      &              99.45        & 99.35                                     &     99.40                                        &                                 99.23         &   92.50                      &    6.80  & 13.30

\\\hline

\begin{tabular}[c]{@{}c@{}}YOLOv5s \\ after freezing 7 layers \cite{sen2024hgr}\end{tabular}                     &       98.90                                  & 98.50                                       &                                          98.70   &         98.95                                 &                      91.16   &     5.90                            & 12.50                                              \\
\hline

\begin{tabular}[c]{@{}c@{}}YOLOv5s \\ after freezing 10 layers \cite{sen2024hgr}\end{tabular}                      &   99.16                                      & 99.08                                       &   99.12                                          &          99.20                                &                91.53         &    5.20                        & 11.95                                                  \\
\hline

\multicolumn{7}{|c|}{\textbf{Our suggested Pruned YOLOv5s (after fine-tuning)}}                                                                                                                                                                                                                                                                                                                                                                                                               \\ 
\hline
\begin{tabular}[c]{@{}c@{}}\textbf{10\% Pruned YOLOv5s}\end{tabular} &     99.70           &      99.60       &    99.65          &     99.50       &   92.50                                                       &      6.2                         & 12                                       \\
\hline

\begin{tabular}[c]{@{}c@{}}\textbf{15\% Pruned YOLOv5s} \end{tabular} &    99.50            &      99.70       &    99.60          &    99.40        &   91.10                                                       &                  5.8                    & 11.5                                \\
\hline

\begin{tabular}[c]{@{}c@{}}\textbf{20\% Pruned YOLOv5s}\end{tabular} & 98.50               &        97.50     &  98.00            &    99.30        &       90.80                                                   &    5.3   & 10.7                                                               \\
\hline

\end{tabular}
}
\end{table*}

\begin{table*}
\centering
\caption{\label{model_comparison_asl_4th} Comparison among various object detection models on ASL dataset.}
\label{model_comparison_asl}
\resizebox*{0.75\textwidth}{!}{
\begin{tabular}{|c|c|c|c|c|c|} 
\hline
\textbf{Models}                                                               & \textbf{mAP50(\%)}    & \textbf{mAP50-95(\%)} & \textbf{Precision(\%)} & \textbf{Recall(\%)}   & \textbf{F-score(\%)}                                     \\ 
\hline
\begin{tabular}[c]{@{}c@{}}Faster-RCNN \\ (Inception V2) \cite{rubin2019hand}\end{tabular}         & 94.20 &  83.45 & 89.60 & 89.20 & 89.40  \\ 

\hline
YOLOv5m                                                                       & 99.10                 & 88.70                 & 98.40                   & 98.80                  & 98.60                  \\ 
\hline
YOLOv5l                                                                       & 99.10       & 88.20          & 98.70                 & 98.50                   & 98.60                                     \\ 
\hline
YOLOv5x                                                                       & 95.6                  & 85.3                  & 95.4                   & 94.9                  & 95.2                   \\ 
\hline
YOLOv7                                                                        &  98.5                     &   88.5                    &          97.6              &        96.8               &     97.2                    
          \\ 

\hline
\hline
\multicolumn{6}{|c|}{\textbf{Proposed channel pruned YOLOv5s (after fine-tuning )}} \\

\hline
\begin{tabular}[c]{@{}c@{}}\textbf{Pruned YOLOv5s} \\ (\textbf{10\% pruned)}\end{tabular}                      &             99.40                            & 89.20                                       &   98.80                                          &            98.60                              &                        98.70                                                                         \\
\hline
\begin{tabular}[c]{@{}c@{}}\textbf{Pruned YOLOv5s} \\  (\textbf{15\% pruned)}\end{tabular}                      &              99.30        & 88.70                                     &     98.60                                        &                                 98.70         &   98.65                       

\\\hline

\begin{tabular}[c]{@{}c@{}}\textbf{Pruned YOLOv5s} \\ (\textbf{20\% pruned)}\end{tabular}                     &       99.30                                  & 87.20                                       &                                          98.40   &         98.90                                 &                      98.65                                                                            \\
\hline

\end{tabular}
}
\end{table*}

\subsection{Building of novel HCI by utilizing fine-tuned pruned YOLOv5s model} \label{HCI_4th}

With the development of technology, multimedia (such as audio and video) has become increasingly important in our daily lives. However, interacting with the devices can be difficult for the physically impaired person. This contribution shows a gesture-controlled HMI to control VLC and Spotify players in real-time. After obtaining the fine-tuned, pruned YOLOv5s model (trained on the `NITR-HGR' dataset), the model's weight file (in .pt format) is further utilized to operate two applications (VLC media player and Spotify player) in real-time scenarios with the correctly classified gesture classes as input commands. The details of our gesture-controlled VLC and Spotify player have been demonstrated in the following subsection.

\begin{figure*}[!ht]
\centering
\includegraphics[width=18cm,height=8.5 cm]{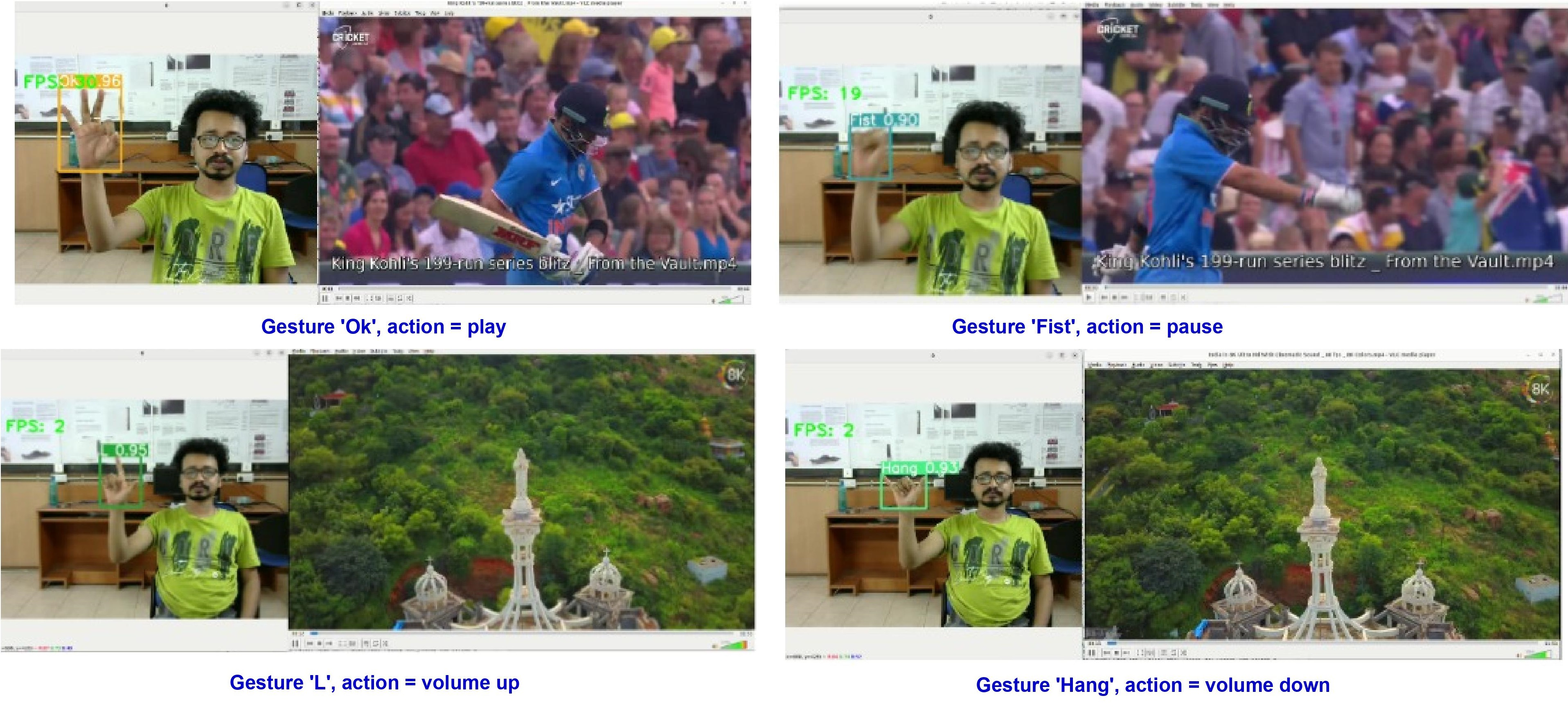}
\caption{\label{vlc_control_4th}Illustration of the gesture-controlled VLC player in real-time. }
\end{figure*}

\begin{figure*}[!ht]
\centering
\includegraphics[width=18cm,height=8.5 cm]{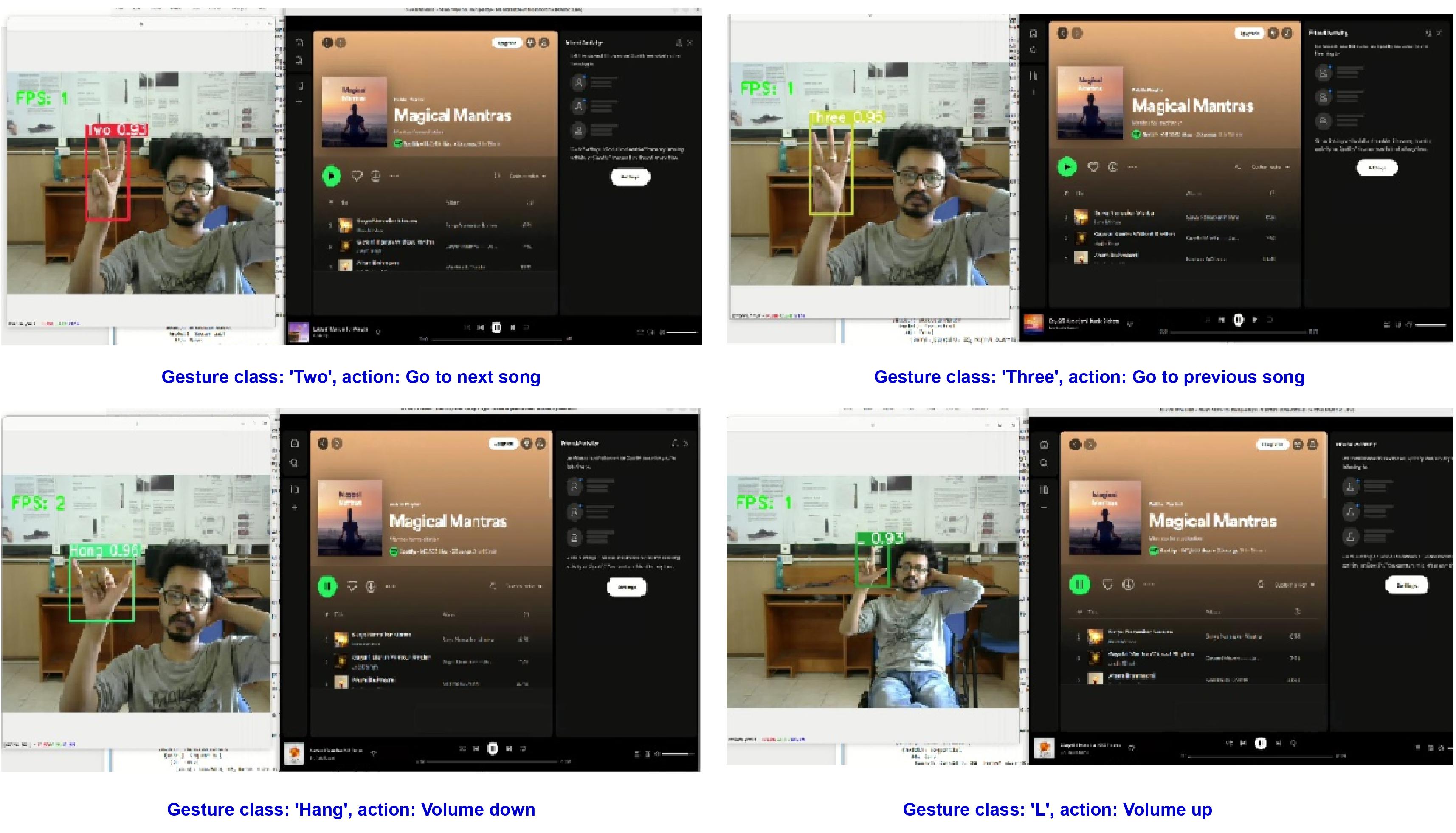}
\caption{\label{spotify_control_4th} Depiction of the gesture-controlled Spotify player in real-time. }
\end{figure*}

\paragraph {\textbf{VLC player and Spotify player control}}
In this work, the `playerctl' \cite{Altdesktop} command-line interface has been utilized to control VLC and Spotify player's functions, such as play, pause, proceed to next video, volume up/down, etc. To operate the VLC player using gesture commands, the keyboard event of the keys is triggered after obtaining the predicted gesture classes. For example, the gesture `Ok' has been set to play a video. Similarly, gestures `Two' and `Three' denote to move the next or previous media in a playlist. 
 We have used predicted gesture classes to trigger each function of the VLC and Spotify player five times to achieve promising outcomes.\\ Table \ref{VLC_audio_performance_4th} displays the performance of the gesture-controlled VLC and Spotify player, including the average response time (in milliseconds), hit and miss counts. Two demos of this gesture-controlled VLC and Spotify player have been shown in Figures \ref{vlc_control_4th} and \ref{spotify_control_4th}.

\begin{table*}[!ht]
\centering
\caption{\label{VLC_audio_performance_4th}Analysis of the real-time performance of gesture-controlled VLC and Spotify player}.
\label{VLC_audio_performance_4th}
\resizebox*{0.75\textwidth}{!}{
\begin{tabular}{|c|c|c|c|c|c|} 
\hline
\multicolumn{6}{|c|}{\textbf{Performance analysis of gesture-controlled VLC player}}                                                                 \\ 
\hline
Action               & Gesture classes & Hit counts & Miss counts & Detection rate (\%) & \begin{tabular}[c]{@{}c@{}}Average Response \\ time (ms)\end{tabular}  \\ 
\hline
Play                 & Ok                     & 5              & 0                 & 100                 & 5.0                                                                    \\ 
\hline
Volume Up            & L                      & 4              & 1                 & 80                  & 8.3                                                                    \\ 
\hline
Volume down          & Hang                   & 4              & 1                 & 80                  & 8.2                                                                    \\ 
\hline
Go to previous video & Three                  & 3              & 2                 & 60                  & 10.3                                                                   \\ 
\hline
Go to next video     & Two                    & 3              & 2                 & 60                  & 12.5                                                                   \\ 
\hline
Pause                & Fist                   & 5              & 0                 & 100                 & 7.3                                                                    \\ 
\hline
\multicolumn{6}{|c|}{\textbf{Performance analysis of gesture-controlled Spotify player}}                                                           \\ 
\hline
Play                 & Ok                     & 5              & 0                 & 100                 & 6.0                                                                    \\ 
\hline
Pause                & Fist                   & 5              & 0                 & 100                 & 7.3                                                                    \\ 
\hline
Volume Up            & L                      & 4              & 1                 & 80                  & 10.9                                                                   \\ 
\hline
Volume Down          & Hang                   & 4              & 1                 & 80                  & 11.5                                                                   \\ 
\hline
Proceed to next track      & Two                    & 3              & 2                 & 60                  & 12.3                                                                   \\ 
\hline
Return to previous song  & Three                  & 3              & 2                 & 60                  & 12.9                                                                   \\
\hline
\end{tabular}
}

\end{table*}

\paragraph{\textbf{Improvement of the robustness of the gesture-controlled VLC and Spotify player}}

While operating VLC/Spotify control with the correctly classified gesture commands, it shows poor results for various functions, such as volume up/down, going to the next, or previous media because of the higher processing of frames sequence in real-time with respect to time. \\
The core functions of the VLC and Spotify players are playing, pausing, volume up/down, and going to the next/previous media/song, etc. Among these functions, playing/pausing are considered as `discrete', and the other remaining are considered as `continuous' functions. A higher frame rate may result in the skipping of desired media from the playlist when the player is being operated using the corresponding gesture commands, which becomes an issue. For instance, if a user indicates the corresponding gesture 'Two' in front of the camera and wishes to proceed to the next song on the gesture-controlled Spotify player, it will bypass the specified song and proceed to multiple others. The user experiences the same result when adjusting the volume using the corresponding gesture commands. To combat these issues, we have delayed the program's execution by a fraction of the time (in seconds). In our experiment, we call the `time sleep ()' function. It reduces the higher frame rate in real-time testing tasks, which benefits the user from triggering continuous functions (such as play/pause, volume up/down, etc) in the Spotify/VLC player with the corresponding gesture commands. By incorporating the sleep() function after each label (gesture class) mapping (demonstrated in Algorithm \ref{up/dowm}), the synchronization is maintained between mapping the right class and activating a continuous function of VLC and Spotify player in real-time. Hence, it brings robustness to the suggested gesture-controlled VLC/Spotify player. The subsequent steps to enhance the robustness of the gesture-controlled VLC and Spotify player are detailed in Algorithm \ref{up/dowm}. In Algorithm \ref{up/dowm}, `Ok', `Two', `Three', `L', `Hang', `Fist' represent the gesture classes in our custom NITR-HGR dataset. 

  \begin{algorithm}[!h]
\caption{Improvement of the gesture-controlled VLC or Spotify player}\label{up/dowm}

    \textbf{INPUT}: Use predicted gesture from fine-tuned pruned YOLOv5s model as input commands. 
    
    \textbf{OUTPUT}: Improved VLC/Spotify player.

\While{True}
{
  Step 1: Capture the frames from the web-camera-captured video streams.\\
  Step 2: Resize the frame to match the input size while training with the baseline YOLOv5s model for gesture detection.\\
  Step 3: Feed the prepossessed frame to the gesture detection model (in our case, fine-tuned pruned YOLOv5s model) to obtain the predicted gesture class. \\
  Step 4: Use the corresponding gesture class to trigger each control of the VLC and Spotify player.

  \uIf{gesture class = ``$Ok$"}{
     
     Play media;
  }
  \uElseIf{gesture class = ``$Fist$"}{
  
   Pause media;
  }

   \uElseIf{gesture class = ``$Two$"}{
       Go to the next media; \\
    Call the sleep () function to decrease the frame rate and prevent the bypass of the desired song;
  }

   \uElseIf{gesture class = ``$Three$"}{
    
    Go to the previous media;\\
    Call the sleep () function to decrease the frame rate;
  }

    \uElseIf{gesture class = ``$L$"}{
         Volume Up; \\
        Call the sleep () function to gradually increase the volume;  }

     \uElseIf{gesture class = ``$Hang$"}{
        Volume Down; \\
        Call the sleep () function to prevent the instant volume drop;
        
          }
  
     \Else
        {
         No action;
        }
  
}
\end{algorithm}

  		
  		
  		
    
            


\section{Discussions}\label{discussion}
In this article, a robust hand gesture recognition system is depicted by using the channel-pruned YOLOv5s model. The suggested model is evaluated on two annotated hand gesture datasets such as the `NITR-HGR' (private) and `ASL' (public). In this contribution, firstly, the YOLOv5s model is chosen for the gesture detection task. Next, the YOLOv5s-based gesture detection model is optimized by using a channel-pruning algorithm to remove the unnecessary channels to obtain a lightweight model for faster inference tasks. It is observed that after pruning, the detection accuracy is slightly reduced, so the pruned model has been further trained using hyper-parameter settings to ensure better detection accuracy. We have demonstrated the performance comparison of the YOLOv5s model for gesture detection after fine-tuning channel-pruned models with different pruning rates varying from (10-20)\% on `NITR-HGR' dataset. It is noticed that mAP, F1-score (\%) have been reduced after increasing the pruning rate, but the detection speed has been increased due to the accuracy-speed trade-off. Our suggested channel-pruned model (with different pruning rates (\%)) has been compared with 12 object detection models and it is observed in Table \ref{model_comparison_nitr}, that our suggested scheme outperformed the other object detection models in terms of precision (\%), recall (\%) and mAP50-95 (\%). We have depicted two correctly classified gesture images (one for a normal person and another for a physically impaired person) using the suggested channel pruned model in Figure \ref{correctly_classified_nitr}. It is observed in Figure \ref{correctly_classified_nitr}, that the detection of the hand portion is promising and occlusion-free. Finally, the fine-tuned, pruned YOLOv5s model is further used to control two multimedia applications (VLC player and Spotify player) using correctly predicted gesture classes as input commands in real-time scenarios. The robustness of the gesture-controlled VLC and Spotify player has been improved. Overall, the improved gesture-controlled VLC and Spotify player system can be suitable to provide a seamless and user-friendly experience.

\section{\label {conclusion} Conclusion}
 This paper presents a robust hand gesture recognition system based on the suggested lightweight channel-pruned YOLOv5s model under complex backgrounds and light-varying conditions. This study also depicts a gesture-controlled HMI to control two applications, including VLC and Spotify player, using predicted gesture class labels in real-time. It is noticed that after introducing the channel-pruned YOLOv5s model, the detection accuracy is slightly decreased compared to the baseline YOLOv5s model. However, the model complexity and the number of parameters have been decreased compared to the original YOLOv5s model with different pruning rates. So to enhance the detection accuracy, the pruned model is fine-tuned with the hyper-parameter settings. In our future work, we will build dynamic hand gesture recognition and utilize the correctly predicted gesture classes to control the multimedia applications in real time with quick response time and higher inference speed. Furthermore, the scope of this work can be expanded by integrating additional modalities, such as eye-gaze tracking and facial expression, to provide a multimodal fusion-based interface.

\section{\label {Data availability} Data availability}
We hereby verify that the dataset will be provided on reasonable request.

\section*{\label {conflict}Conflict of interest} 
We declare that we have no conflict of interest.


%

\bibliographystyle{IEEEtran}
\bibliography{bibfile}

\begin{thebibliography}{10}
\providecommand{\url}[1]{#1}
\csname url@samestyle\endcsname
\providecommand{\newblock}{\relax}
\providecommand{\bibinfo}[2]{#2}
\providecommand{\BIBentrySTDinterwordspacing}{\spaceskip=0pt\relax}
\providecommand{\BIBentryALTinterwordstretchfactor}{4}
\providecommand{\BIBentryALTinterwordspacing}{\spaceskip=\fontdimen2\font plus
\BIBentryALTinterwordstretchfactor\fontdimen3\font minus
  \fontdimen4\font\relax}
\providecommand{\BIBforeignlanguage}[2]{{%
\expandafter\ifx\csname l@#1\endcsname\relax
\typeout{** WARNING: IEEEtran.bst: No hyphenation pattern has been}%
\typeout{** loaded for the language `#1'. Using the pattern for}%
\typeout{** the default language instead.}%
\else
\language=\csname l@#1\endcsname
\fi
#2}}
\providecommand{\BIBdecl}{\relax}
\BIBdecl

\bibitem{berezhnoy2018hand}
V.~Berezhnoy, D.~Popov, I.~Afanasyev, and N.~Mavridis, ``The hand-gesture-based
  control interface with wearable glove system.'' in \emph{ICINCO (2)}, 2018,
  pp. 458--465.

\bibitem{abhishek2016glove}
K.~S. Abhishek, L.~C.~F. Qubeley, and D.~Ho, ``Glove-based hand gesture
  recognition sign language translator using capacitive touch sensor,'' in
  \emph{2016 IEEE International Conference on Electron Devices and Solid-State
  Circuits (EDSSC)}.\hskip 1em plus 0.5em minus 0.4em\relax IEEE, 2016, pp.
  334--337.

\bibitem{singha2018dynamic}
J.~Singha, A.~Roy, and R.~H. Laskar, ``Dynamic hand gesture recognition using
  vision-based approach for human--computer interaction,'' \emph{Neural
  Computing and Applications}, vol.~29, no.~4, pp. 1129--1141, 2018.

\bibitem{al2022structured}
F.~Al~Farid, N.~Hashim, J.~Abdullah, M.~R. Bhuiyan, W.~N. Shahida Mohd~Isa,
  J.~Uddin, M.~A. Haque, and M.~N. Husen, ``A structured and methodological
  review on vision-based hand gesture recognition system,'' \emph{Journal of
  Imaging}, vol.~8, no.~6, p. 153, 2022.

\bibitem{elmezain2008hidden}
M.~Elmezain, A.~Al-Hamadi, J.~Appenrodt, and B.~Michaelis, ``A hidden markov
  model-based continuous gesture recognition system for hand motion
  trajectory,'' in \emph{2008 19th international conference on pattern
  recognition}.\hskip 1em plus 0.5em minus 0.4em\relax IEEE, 2008, pp. 1--4.

\bibitem{yang2012dynamic}
Z.~Yang, Y.~Li, W.~Chen, and Y.~Zheng, ``Dynamic hand gesture recognition using
  hidden markov models,'' in \emph{2012 7th International Conference on
  Computer Science \& Education (ICCSE)}.\hskip 1em plus 0.5em minus
  0.4em\relax IEEE, 2012, pp. 360--365.

\bibitem{huang2009vision}
D.-Y. Huang, W.-C. Hu, and S.-H. Chang, ``Vision-based hand gesture recognition
  using pca+ gabor filters and svm,'' in \emph{2009 fifth international
  conference on intelligent information hiding and multimedia signal
  processing}.\hskip 1em plus 0.5em minus 0.4em\relax IEEE, 2009, pp. 1--4.

\bibitem{rahman2013hand}
M.~H. Rahman, J.~Afrin \emph{et~al.}, ``Hand gesture recognition using
  multiclass support vector machine,'' \emph{International Journal of Computer
  Applications}, vol.~74, no.~1, pp. 39--43, 2013.

\bibitem{yingxin2016robust}
X.~Yingxin, L.~Jinghua, W.~Lichun, and K.~Dehui, ``A robust hand gesture
  recognition method via convolutional neural network,'' in \emph{2016 6th
  international conference on digital home (ICDH)}.\hskip 1em plus 0.5em minus
  0.4em\relax IEEE, 2016, pp. 64--67.

\bibitem{SHARMA2021115657}
\BIBentryALTinterwordspacing
S.~Sharma and S.~Singh, ``Vision-based hand gesture recognition using deep
  learning for the interpretation of sign language,'' \emph{Expert Systems with
  Applications}, vol. 182, p. 115657, 2021. [Online]. Available:
  \url{https://www.sciencedirect.com/science/article/pii/S0957417421010484}
\BIBentrySTDinterwordspacing

\bibitem{sen2022novel}
A.~Sen, T.~K. Mishra, and R.~Dash, ``A novel hand gesture detection and
  recognition system based on ensemble-based convolutional neural network,''
  \emph{Multimedia Tools and Applications}, pp. 1--24, 2022.

\bibitem{redmon2016you}
J.~Redmon, S.~Divvala, R.~Girshick, and A.~Farhadi, ``You only look once:
  Unified, real-time object detection,'' in \emph{Proceedings of the IEEE
  conference on computer vision and pattern recognition}, 2016, pp. 779--788.

\bibitem{girshick2015fast}
R.~Girshick, ``Fast r-cnn,'' in \emph{Proceedings of the IEEE international
  conference on computer vision}, 2015, pp. 1440--1448.

\bibitem{bose2020efficient}
S.~R. Bose and V.~S. Kumar, ``Efficient inception v2 based deep convolutional
  neural network for real-time hand action recognition,'' \emph{IET Image
  Processing}, vol.~14, no.~4, pp. 688--696, 2020.

\bibitem{yu2019hand}
X.~Yu and Y.~Yuan, ``Hand gesture recognition based on faster-rcnn deep
  learning.'' \emph{Journal of Computers}, vol.~14, no.~2, pp. 101--110, 2019.

\bibitem{sen2024hgr}
A.~Sen, S.~Dombe, T.~K. Mishra, and R.~Dash, ``Hgr-fyolo: a robust hand gesture
  recognition system for the normal and physically impaired person using frozen
  yolov5,'' \emph{Multimedia Tools and Applications}, pp. 1--19, 2024.

\bibitem{rautaray2010novel}
S.~S. Rautaray and A.~Agrawal, ``A novel human computer interface based on hand
  gesture recognition using computer vision techniques,'' in \emph{Proceedings
  of the First International Conference on Intelligent Interactive Technologies
  and Multimedia}, 2010, pp. 292--296.

\bibitem{sen2023deep}
A.~Sen, T.~K. Mishra, and R.~Dash, ``Deep learning-based hand gesture
  recognition system and design of a human--machine interface,'' \emph{Neural
  Processing Letters}, pp. 1--28, 2023.

\bibitem{everingham2009the}
M.~Everingham, L.~V. Gool, C.~K.~I. Williams, J.~Winn, and A.~Zisserman, ``The
  pascal visual object classes (voc) challenge,'' \emph{International Journal
  of Computer Vision}, vol.~88, pp. 303--308, September 2009.

\bibitem{lin2014microsoft}
T.-Y. Lin, M.~Maire, S.~Belongie, J.~Hays, P.~Perona, D.~Ramanan,
  P.~Doll{\'a}r, and C.~L. Zitnick, ``Microsoft coco: Common objects in
  context,'' in \emph{Computer Vision--ECCV 2014: 13th European Conference,
  Zurich, Switzerland, September 6-12, 2014, Proceedings, Part V 13}.\hskip 1em
  plus 0.5em minus 0.4em\relax Springer, 2014, pp. 740--755.

\bibitem{he2015spatial}
K.~He, X.~Zhang, S.~Ren, and J.~Sun, ``Spatial pyramid pooling in deep
  convolutional networks for visual recognition,'' \emph{IEEE transactions on
  pattern analysis and machine intelligence}, vol.~37, no.~9, pp. 1904--1916,
  2015.

\bibitem{liu2017learning}
Z.~Liu, J.~Li, Z.~Shen, G.~Huang, S.~Yan, and C.~Zhang, ``Learning efficient
  convolutional networks through network slimming,'' in \emph{Proceedings of
  the IEEE international conference on computer vision}, 2017, pp. 2736--2744.

\bibitem{asl-recognition-uhnrr_dataset}
\BIBentryALTinterwordspacing
A.~Recognition, ``Asl recognition dataset,'' \url{
  https://universe.roboflow.com/asl-recognition/asl-recognition-uhnrr }, nov
  2022, visited on 2023-04-14. [Online]. Available:
  \url{https://universe.roboflow.com/asl-recognition/asl-recognition-uhnrr}
\BIBentrySTDinterwordspacing

\bibitem{rubin2019hand}
S.~Rubin~Bose and V.~Sathiesh~Kumar, ``Hand gesture recognition using faster
  r-cnn inception v2 model.'' in \emph{AIR}, 2019, pp. 19--1.

\bibitem{mujahid2021real}
A.~Mujahid, M.~J. Awan, A.~Yasin, M.~A. Mohammed, R.~Dama{\v{s}}evi{\v{c}}ius,
  R.~Maskeli{\=u}nas, and K.~H. Abdulkareem, ``Real-time hand gesture
  recognition based on deep learning yolov3 model,'' \emph{Applied Sciences},
  vol.~11, no.~9, p. 4164, 2021.

\bibitem{hu2022gesture}
D.~Hu, J.~Zhu, J.~Liu, J.~Wang, and X.~Zhang, ``Gesture recognition based on
  modified yolov5s,'' \emph{IET Image Processing}, vol.~16, no.~8, pp.
  2124--2132, 2022.

\bibitem{Altdesktop}
\BIBentryALTinterwordspacing
p.~Altdesktop, ``Altdesktop/playerctl: Media player command-line controller for
  vlc, mpv, rhythmbox, web browsers, cmus, mpd, spotify and others.'' [Online].
  Available: \url{https://github.com/altdesktop/playerctl}
\BIBentrySTDinterwordspacing

\end{thebibliography}

\end{document}